\documentclass[pdflatex,sn-mathphys-num]{sn-jnl}

\usepackage{graphicx}%
\usepackage{multirow}%
\usepackage{amsmath,amssymb,amsfonts}%
\usepackage{amsthm}%
\usepackage{mathrsfs}%
\usepackage[title]{appendix}%
\usepackage{xcolor}%
\usepackage{textcomp}%
\usepackage{manyfoot}%
\usepackage{booktabs}%
\usepackage{algorithm}%
\usepackage{algorithmicx}%
\usepackage{algpseudocode}%
\usepackage{listings}%

\begin{document}

\title{
XiCAD: Camera Activation Detection 

in the Da Vinci Xi User Interface
}


\author*[1]{\fnm{Alexander C.} \sur{Jenke}}\email{alexander.jenke@nct-dresden.de}

\author[1]{\fnm{Gregor} \sur{Just}}

\author[1]{\fnm{Claas} \sur{de Boer}}

\author[2, 3]{\fnm{Martin}, \sur{Wagner}}

\author[1, 3]{\fnm{Sebastian} \sur{Bodenstedt}}

\author[1, 3]{\fnm{Stefanie} \sur{Speidel}}

\affil[1]{Department of Translational Surgical Oncology, National Center for Tumor Diseases (NCT), NCT/UCC Dresden, a partnership between DKFZ, Faculty of Medicine and University Hospital Carl Gustav Carus, TUD Dresden University of Technology, and Helmholtz-Zentrum Dresden-Rossendorf (HZDR), Germany}
\affil[2]{Department of Visceral, Thoracic and Vascular Surgery, Faculty of Medicine and University Hospital Carl Gustav Carus, TUD, Germany}
\affil[3]{Cluster of Excellence-CeTI, TUD, Germany}

\abstract{
\textbf{Purpose:} 
Robot-assisted minimally invasive surgery relies on endoscopic video as the sole intraoperative visual feedback. 
The DaVinci Xi system overlays a graphical user interface (UI) that indicates the state of each robotic arm, including the activation of the endoscope arm. 
Detecting this activation provides valuable metadata such as camera movement information, which can support downstream surgical data science tasks including tool tracking, skill assessment, or camera control automation.

\textbf{Methods:} 
We developed a lightweight pipeline based on a ResNet18 convolutional neural network to automatically identify the position of the camera tile and its activation state within the DaVinci Xi UI.
The model was fine-tuned on manually annotated data from the SurgToolLoc dataset and evaluated across three public datasets comprising over 70,000 frames.

\textbf{Results:} 
The model achieved F1-scores between 0.993 and 1.000 for the binary detection of active cameras and correctly localized the camera tile in all cases without false multiple-camera detections.

\textbf{Conclusion:} 
The proposed pipeline enables reliable, real-time extraction of camera activation metadata from surgical videos, facilitating automated preprocessing and analysis for diverse downstream applications. 
All code, trained models, and annotations are publicly available.
}


\maketitle

\section{Introduction}
Robot-assisted minimally invasive surgery (RAMIS) relies on endoscopic video as the sole source of intraoperative visual feedback.
The graphical user interface (UI) of robotic systems is typically overlaid on these videos and provides information about attached instruments and their operational states.
While surgical data science commonly analyzes endoscopic video streams to study surgical workflow, skill, or anatomy, the embedded UI is rarely utilized, despite offering valuable and easily accessible metadata such as instrument and camera activation states.

In the DaVinci Xi system, activation of the camera arm directly indicates camera movement, as the endoscope can only be repositioned while the arm is actively controlled.
This movement information is highly relevant for multiple downstream tasks.
For simultaneous localization and mapping (SLAM), it enables the selection of scenes with either moving or stationary cameras. In surgical skill analysis, the extent of camera motion can serve as an indicator of camera handling proficiency. When analyzing individual frames, motion often causes blur, and such frames can be filtered accordingly. In tool tracking, distinguishing between instrument and camera motion is essential, as camera movement distorts the measured path length and range of motion. Finally, for training robot-assisted camera control, learning when and how to move the camera directly corresponds to the activation of the camera arm in RAMIS.

As this information is inherently available within recorded surgical videos containing the da Vinci Xi UI, automatic extraction of this metadata would significantly facilitate their reuse in downstream tasks.
In this work, we present a lightweight pipeline that automatically detects the position of the camera tile and its activation state in the da Vinci Xi UI.
Additionally, we release the manually created annotations used for training and evaluation across three publicly available datasets. 

The complete pipeline, model, and annotations are available at \url{https://gitlab.com/nct_tso_public/xicad}.

\section{Method}

    \begin{figure}[htbp!]
        \centering
        \includegraphics[width=.8\linewidth]{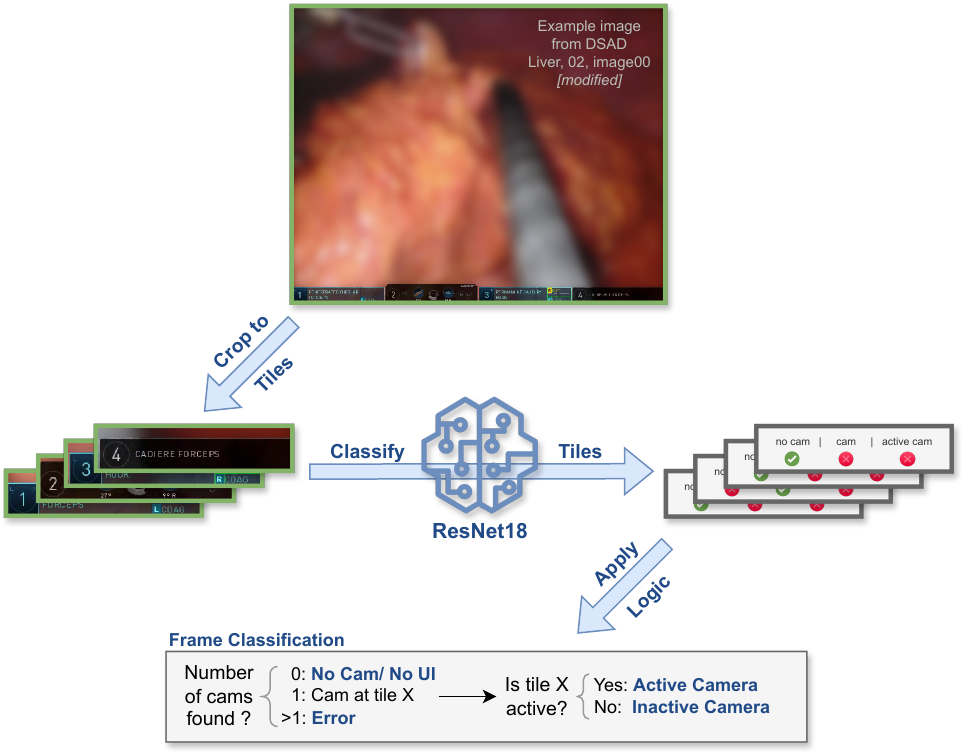}
        \caption{
        Overview of the proposed pipeline for detecting the camera tile and its activation state in the DaVinci Xi user interface. 
        Four tiles are cropped out of the frame, each is classified by a fine-tuned ResNet18 as no camera, inactive camera, or active camera. 
        The tile-level predictions are then combined through simple logic to yield the final frame-level camera activation and position.
        }
        \label{fig:pipeline}
    \end{figure}

\subsection{DaVinci Xi UI}
    The DaVinci Xi UI is overlaid on the endoscopic video stream.
    It mainly consists of four semi-transparent rectangular tiles shown continuously at the bottom of the frame (see example in Fig. \ref{fig:pipeline}).
    Other UI elements like virtual pointers, arm popups, off-screen indicators or system status messages will not be used in this work and are therefore not further introduced. 

    Each tile represents the state of one of the four robot arms, which hold up to three instruments and the endoscope.
    The tiles are numbered one to four identifying which arm they represent. 
    The numbers are not necessarily ordered, but in the used datasets this is generally the case.
    Furthermore, the tiles either show the instrument name and activation status for instruments or icons describing table orientation, endoscope horizon, zoom levels, and view angels for the endoscope.
    A light blue highlighting of the tile indicates which arms are currently being controlled by the surgeon. 
    At any time this can either be two of the three instruments or just the endoscope.

    The endoscope tile only becomes highlighted when the surgeon actively presses and holds down the endoscope control foot pedal. 
    Consequently, the camera is not activated in idle situations, which allows us to assume that the camera is very likely being moved when the tile is highlighted.

    To extract the camera activity from the UI, we first need to detect which of the four tiles shows the endoscope information and subsequently check whether the tile is active or not. 
    Firstly, we remove any black border around the endoscope image, resulting in a 5:4 aspect ratio, and rescale the image to $640\times521$ pixels.
    Next, we crop the image, yielding four cropped tile images. 
    To accommodate for slight shifts in the UI placement due to padding or compression in the recording setups, each tile covers an additional area of four pixels around the expected position, resulting in $168\times28$ pixels for each tile. 
    Each tile is then processed separately by the neural network.

\subsection{Neural network}
    The neural network consists of a ResNet18\cite{resnet18} pretrained on ImageNet\cite{imagenet} and fine-tuned on the given task.
    The classification layer was replaced to output three classes: \textit{no camera}, an \textit{inactive camera} and an \textit{active camera}.
    The small model takes the input image of $168\times28$ pixels and predicts a scalar for all three classes, which are further reduced to the most likely class using argmax during inference. 

    As the four tiles of the UI are processed separately, the results need further post-processing during inference to classify the whole frame. 
    Based on the output of all four tiles the frame is classified into one of the four classes \textit{no camera}, \textit{one inactive camera},\textit{one active camera}, and \textit{too many cameras}.
    Additionally, the position ID of the corresponding tile is returned if exactly one camera is found, being labeled T1 to T4 from left to right. 

\subsection{Datasets}
    The neural network was fine-tuned on 7699 manually labeled tiles extracted from 2265 videos from the SurgToolLoc\cite{SurgToolLoc} dataset. 
    The distribution of classes per position ID are shown in Table \ref{tab:classperpos}.

    \begin{table}[htpb!]
        \centering
        \begin{tabular}{l|c|c|c|c||c}
            tile shows & T1 & T2 & T3 & T4 & all positions \\
             \hline
            active camera & 0 & 86 & 27 & 0 & 113 \\
            inactive camera & 0 & 510 & 258 & 0 & 768 \\
            not a camera & 1883 & 289 & 1437 & 3209 & 6818 \\
        \end{tabular}
        \caption{Number of training tiles per class (active camera, inactive camera, not a camera) for each UI tile position (T1–T4) in the SurgToolLoc dataset, including the total number across all positions.}
        \label{tab:classperpos}
    \end{table}

    The performance was evaluated on three datasets: SurgToolLoc, HSDB Instrument\cite{hsdb}, and DSAD\cite{dsad}.
    These were manually annotated framewise, providing, if existing, the position of the camera tile and its activation.
    For the SurgToolLoc 6246 videos are used, which are not in the training data.
    The distribution of the classes is shown in Table \ref{tab:testclasses}.

    \begin{table}[htbp!]
        \centering
        \begin{tabular}{l|c|c|c|c|c|c|c|c|c||c}
            & no & \multicolumn{4}{c|}{inactive camera} & \multicolumn{4}{c||}{active camera} & number of \\
            dataset & camera & T1 & T2 & T3 & T4 & T1 & T2 & T3 & T4 & frames \\
            \hline
            SurgToolLoc & 8353 & 0 & 1469 & 14791 & 0 & 0 & 47 & 323 & 0 & 24983 \\
            HSDB Instrument & 16262 & 10 & 17574 & 781 & 0 & 0 & 949 & 0 & 0 & 35576 \\
            DSAD & 6 & 0 & 251 & 12345 & 51 & 0 & 0 & 542 & 0 & 13195
        \end{tabular}
        \caption{Number of annotated frames per class and tile position across the three test datasets, including the total number of frames.}
        \label{tab:testclasses}
    \end{table}

\subsection{Experiment Setup}
    After extracting the annotated tiles from the train dataset, the neutral network was trained for 100 epochs using a batch size of 128.
    The first block of the ResNet18 was frozen, and the remaining parameters updated using the SGD optimizer\cite{sgd} with a momentum of 0.9.
    The learning rate was scheduled using a OneCycleLR scheduler\cite{onecyclelr} reaching a maximum learning rate of 1e-3.
    Binary cross entropy was used as loss.
    Due to the label imbalance a positive weighting was applied.
    To counteract occasional cases of exploding gradients, gradient norm clipping was applied, limiting the gradient norm to a maximum of 5.

    During training the model performance was evaluated using the F1-Score on the single tiles, averaging the classes without weights. 
    The final model is evaluated on the test sets using F1-Score on the single tiles.
    Further, the final model is evaluated on the test sets using the post processed model output, classifying the whole image over all four tiles, regarding the binary decision of active camera versus inactive or non-existing.
    Additionally, we evaluate the subset of frames ignoring all frames without a camera tile, either not providing a UI or not having a camera attached. 

    The implementation was done using Python3.11 and the PyTorch-Framework\cite{pytorch}.
    For training and validation an Nvidia RTX A5000 GPU was used.
    The used GPU RAM during single frame inference did not exceed 339 MB.
    
\section{Results}
    \begin{figure}[htbp!]
        \centering
        \includegraphics[width=0.45\linewidth]{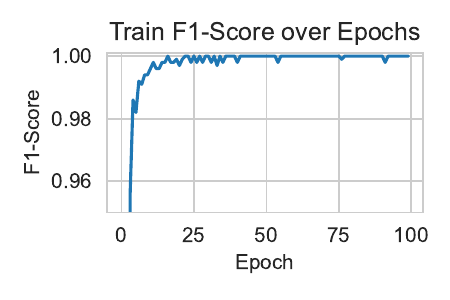}
        \caption{Macro-averaged F1-score across all three tile classes (no camera, inactive camera, active camera) during training over 100 epochs, showing convergence and achievement of perfect classification.}
        \label{fig:f1_train}
    \end{figure}
    
    As shown in figure \ref{fig:f1_train} the model achieved a perfect F1-Score of 1 during training, correctly classifying each tile without a single error.
    The F1-Score of 1 was achieved for the first time in epoch 22.
    The training was continued until the 100th epoch, further reducing the loss. 

    \begin{table}[htbp!]
        \centering
        \begin{tabular}{l|c|c|c|c|c|c|c|c}
            & \multicolumn{4}{c|}{including frames without UI} & \multicolumn{4}{c}{excluding frames without UI}\\
            dataset & Acc. & Prec. & Rec. & F1 & Acc. & Prec. & Rec. & F1 \\
            \hline
            SurgToolLoc     & 1.000 & 0.987 & 1.000 & 0.993 & 1.000 & 1.000 & 1.000 & 1.000 \\
            DSAD            & 1.000 & 1.000 & 0.994 & 0.997 & 1.000 & 1.000 & 0.994 & 0.997 \\
            HSDB Instrument & 0.999 & 1.000 & 0.950 & 0.975 & 0.998 & 1.000 & 0.950 & 0.975\\
        \end{tabular}
        \caption{Binary camera activation detection performance on test datasets, including and excluding frames containing a camera tile.}
        \label{tab:testmetrics}
    \end{table}
    
    The final model achieves an F1-Score of 0.995 on all annotated tiles of all three test sets.
    As shown in table \ref{tab:testmetrics} the binary decision - between active versus inactive or non-existent cameras - achieves an F1-Score of 0.993 - 0.997 for all datasets.
    When evaluated only on frames with a camera tile present the F1-Score of the SurgToolLoc dataset reaches 1.000.
    
    \begin{figure}[htbp!]
        \centering
        \includegraphics[width=\linewidth]{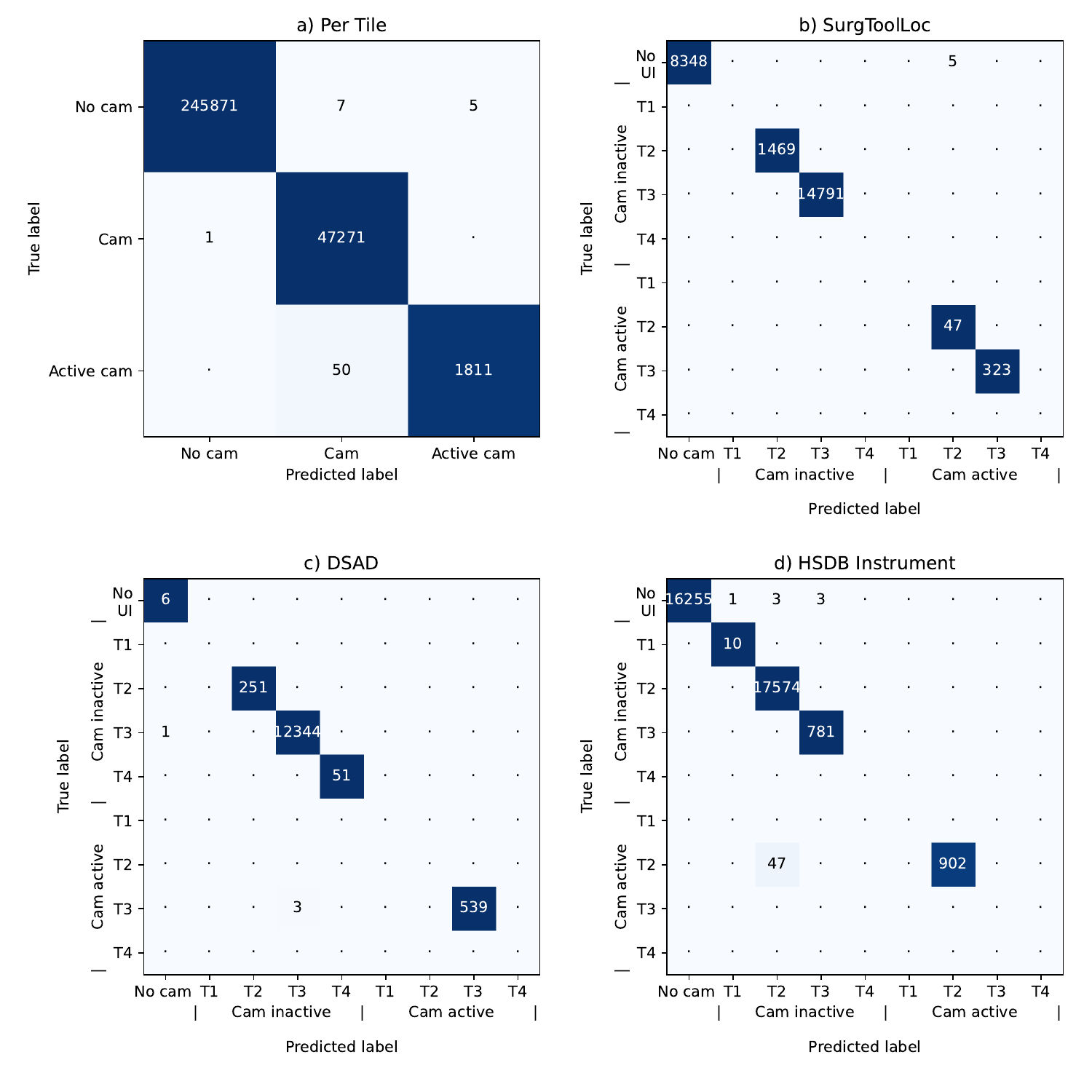}
        \caption{Confusion matrices for tile-level (a) and frame-level (b–d) predictions on the test datasets. Absolute counts are shown, with color indicating percentage normalized over true labels.}
        \label{fig:cms}
    \end{figure}

    As shown in figure \ref{fig:cms} the model detects five active cameras where no camera should be found in the SurgToolLoc test set. 
    In the DSAD dataset the detection of one inactive camera is missed, three active cameras are wrongfully classified as inactive. 
    In the HSDB Instrument dataset 47 active cameras are classified as inactive and in seven cases an inactive camera is detected where no camera should be found. 

    In no case more than one camera was found in a frame. 
    Therefore, no error case needs to be accounted for during evaluation. 
    
    Single-frame inference, where the four tiles of one frame are processed in parallel, achieved an average speed of 100 FPS.

\section{Discussion}
    While the processing of each tile separately allowed for training with fewer annotations, it posed the risk of negatively impacting the detection of the correct number of camera tiles. 
    But as the evaluation showed, this was not the case, as in no case two or more cameras were detected. Further figure \ref{fig:cms} also shows that in no case the position of the detected camera tile was wrong. 
    If a camera was detected, the correct position was found. 

    While the train dataset did not contain a camera at position T1 or T4, all cameras at these positions in the DSAD and HSDB Instrument dataset were correctly detected. 
    As no active camera at positions T1 or T4 are available in all three datasets, the detection of an active camera can not be evaluated.
    Nevertheless, it is reasonable to assume that the model would correctly detect these cases, given its demonstrated ability to generalize to inactive cameras at the same positions.
    
    The prevalence of the camera being placed at position T2 rather than T3, is reflected in the performances of the DSAD and HSDB Instrument dataset. 
    The DSAD dataset, which contains active cameras only at position T2 performs better than the HSDB Instrument dataset, which contains active cameras only at position T3.
    Nevertheless, the difference in F1-Score is only 0.022, and still very good.

    Especially the perfect precision of 1.000 in all three datasets, if we expect a UI to be present, needs to be highlighted.
    The recall can further be improved when processing complete videos rather than single frames. 
    Single erroneous frames can be filtered when smoothing over time, as the activation of the camera tile is a manual action triggered by a foot pedal and can be expected to last a time exceeding the duration of a single frame. 

    The model’s single-frame inference speed of 100 FPS exceeds real-time requirements for endoscopic video. For offline analysis, batching multiple frames or tile sets on the GPU can further accelerate processing, enabling large-scale extraction of camera activation metadata with minimal computational overhead.

\section{Conclusion}

    In this paper, we presented a pipeline to detect the position and activation status of the camera tile in the DaVinci Xi UI, demonstrating strong generalization by achieving an accuracy of at least 0.999 on three different test datasets, while being trained only on the training set of one of them.
    The pipeline can be applied to any video stream capturing the DaVinci Xi UI without the need for any preprocessing, cropping borders or exactly locating the UI.
    Its compact architecture enables real-time deployment or can run even faster using batched processing in an offline setup.

    This easy to use pipeline and almost perfect scores on all evaluation metrics facilitates automatic metadata extraction for downstream surgical video analysis tasks such as phase recognition and skill assessment, or robot-assisted camera control.

\backmatter
\newpage
\section*{Declarations}
\textbf{Conflict of interest} All authors declare no conflict of interest \\
\textbf{Ethics approval and consent to participate} Not applicable\\
\textbf{Consent for publication} Not applicable\\
\textbf{Materials availability} Not applicable\\
\textbf{Code \& Data availability} Code, model, and annotations can be found online. The used datasets are publicly available. \\
\textbf{Author contribution} Conceptualization: ACJ, GJ, CdB, MW; Methodology: ACJ GJ, CdB; Data annotation: ACJ; Software: ACJ; Validation: ACJ;  Visualization: ACJ; Writing - Original Draft: ACJ; Writing - Review \& Editing: ACJ, GJ, CdB, MW, SB, SS; Supervision: SB, SS. \\
\textbf{Use of LLMs} The authors used ChatGPT (OpenAI) solely for linguistic refinement; all content and conclusions were produced by the authors.

\section*{Acknowledgements}
Funded or Co-funded by the European Union through NEARDATA under grant agreement ID 101092644.

This work is supported by the project ``Next Generation AI Computing (gAIn)'', funded by the Bavarian Ministry of Science and the Arts and the Saxon Ministry for Science, Culture, and Tourism.

The authors acknowledge the financial support by the Federal Ministry of Research, Technology and Space of Germany in the programme of “DigiLeistDAT”. Joint project SurgicalAIHubGermany, project identification number: 02K23A112.

Funded by the German Research Foundation (DFG, Deutsche Forschungsgemeinschaft) as part of Germany’s Excellence Strategy – EXC 2050/1 – Project ID 390696704 – Cluster of Excellence “Centre for Tactile Internet with Human-in-the-Loop” (CeTI) of Technische Universität Dresden.

The authors acknowledge the financial support by the  Federal Ministry of Research, Technology and Space of Germany in the programme of “Souverän. Digital. Vernetzt.”. Joint project 6G-life, project identification number: 16KISK001K


\bibliography{main}

\end{document}